\documentclass[11pt]{article}
\usepackage[utf8]{inputenc}
\usepackage[T1]{fontenc}
\usepackage{lmodern}
\usepackage{geometry}
\usepackage{graphicx}
\usepackage{booktabs}
\usepackage{longtable}
\usepackage{calc}
\usepackage{array}
\usepackage{multirow}
\usepackage{caption}
\usepackage{needspace}
\newcommand{\minorheading}[1]{\par\Needspace{5\baselineskip}\medskip\noindent\textbf{#1}\par\nobreak\smallskip}
\newcommand{\figurenote}[1]{\par\vspace{0.35em}\noindent\begin{minipage}{0.92\linewidth}\raggedright\footnotesize\textit{Note.}~#1\end{minipage}\par\vspace{0.75em}}
\usepackage{float}
\usepackage{amsmath,amssymb}
\usepackage{hyperref}
\hypersetup{colorlinks=true,linkcolor=blue,citecolor=blue,urlcolor=blue}
\usepackage{microtype}
\usepackage{enumitem}
\providecommand{\tightlist}{\setlength{\itemsep}{0pt}\setlength{\parskip}{0pt}}
\setlist{itemsep=2pt,topsep=4pt}
\title{Bounded Sovereignty and the Control Tax: Pricing AI Oversight When the Deployer Does Not Own the Model}
\author{Zhen Wen Lim\thanks{AI safety and governance advisor with professional experience in financial services and payments. The views expressed are the author's own.}}
\date{July 2026}

\begin{document}
\maketitle

\begin{abstract}
AI control research asks how to deploy a model safely even when the model may be misaligned, but many control protocols assume that the deployer can instrument the model and its surrounding pipeline. That assumption is often too strong for regulated organisations that consume frontier models through APIs or managed endpoints: they may control the business process, but not the model weights, serving infrastructure, internal traces, update process, or full interaction logs. This paper introduces bounded sovereignty: partial technical and contractual access across the data, model, infrastructure, and interaction layers of the AI stack. The argument is that these access conditions determine which control protocols can be executed in practice. The paper makes five contributions: a four-layer access typology for deployment-time AI control; a protocol-by-layer matrix mapping common control protocols to the access they require; the concept of sovereignty discount cost, meaning the part of the control tax spent substituting for missing access through contracts, architecture, audit, vendor assurance, residual risk, or reduced system scope; a synthetic access-ablation experiment over 1.35 million synthetic case simulations; and an anonymised national-payments-infrastructure scenario in an AI-importing jurisdiction. The experiment is not real-world payment-system evidence, but a construct-validity exercise showing how logging, gateway access, trace access, model-version control, vendor support, tool scope, thresholds, audit budgets, and agent behaviour change prevention, diagnosis, usefulness, audit burden, and cost. The main finding is that access assumptions are safety assumptions: complete logs help a deployer understand what happened, a pre-execution gateway helps it stop risky actions before they occur, trace access and model-version control improve post-incident diagnosis, and scope restriction can improve safety while reducing usefulness. Control protocols proposed as general safety solutions should therefore state their access assumptions explicitly.
\end{abstract}


\section{Introduction}\label{introduction}

\subsection{Control is rationed before it is engineered}\label{control-is-rationed-before-it-is-engineered}

AI safety work often starts from a technical question: if a model may be
misaligned, what deployment protocol can keep its actions safe? This paper
starts one step earlier. Before a deployer can run a control protocol, it
must be able to observe, log, route, interrupt, or audit the relevant parts
of the system. Many deployers do not have that access.

This matters because access to the AI stack is unevenly distributed. In
December 2025, seven countries signed the Pax Silica Declaration in
Washington, a US-led framework for secure and resilient AI-related supply
chains. By late June 2026, the initiative had expanded to twenty-four
signatories, while thirty-five countries and regions signed the related
Joint Statement on AI Opportunity at the second Pax Silica Summit
(Australian Department of Industry, Science and Resources, 2025; Japanese
Ministry of Foreign Affairs, 2026; European Commission, 2026). In the same
period, assessments of global compute distribution reported that Africa
hosts under one percent of global data-center capacity while holding roughly
eighteen percent of the world's population, and that much of South Asia,
Southeast Asia, and Latin America depends on externally owned and operated
AI infrastructure (CSIS, 2025; Brookings Institution, 2026).

These facts are usually discussed as industrial policy or geopolitics. Here
they are treated as a boundary condition for AI control. If a regulated
institution uses a vendor model through an API, it may not control the model
weights, the serving stack, the update process, or the internal traces. It
may still be accountable for the system's behaviour, but it cannot use every
control protocol described in the AI safety literature. The question is not
whether sovereign AI is desirable as national policy. The narrower question
is: when the deployer does not own the model, what control remains possible?

\subsection{The two-layer control problem}\label{the-two-layer-control-problem}

The AI control agenda asks how to keep deployment safe even if a model may
be misaligned (Greenblatt et al., 2023). Canonical protocols include trusted
monitoring, defer-to-trusted routing, resampling-based intervention (Bhatt
et al., 2025), and human auditing of suspicious actions. Recent papers
stress-test these protocols adversarially (Griffin et al., 2024; Schaeffer
et al., 2026; Terekhov et al., 2026), and Terekhov et al.~(2025) price the
\emph{control tax}: the operational and financial cost of adding control to
AI pipelines.

This paper argues that control has two layers:

\begin{itemize}
\tightlist
\item
  \textbf{Layer 1: protocol effectiveness.} If the deployer has the needed
  access, does the protocol reduce unsafe behaviour?
\item
  \textbf{Layer 2: access feasibility.} Does the deployer actually have the
  logs, traces, routing control, model-version control, infrastructure
  visibility, or contractual rights that the protocol assumes?
\end{itemize}

Most AI-control work focuses on Layer 1. This paper focuses on Layer 2. By
\emph{sovereignty}, it does not mean territorial sovereignty. It means
technical and contractual access to the layers that control protocols rely
on: data, model, infrastructure, and interaction surfaces. For many
real-world regulated deployers, especially in AI-importing jurisdictions,
that access is partial.

\subsection{Case motivation, contributions, and paper structure}\label{why-financial-infrastructure-is-a-useful-test-case}

Financial infrastructure is a useful setting for this question because the
risks are concrete, the audit expectations are high, and the accountability
chain is clear. A national payments infrastructure operator may use a
vendor-supplied foundation model to support a transaction-adjacent process,
such as dispute triage or investigation drafting. The domestic regulator
may expect validation, monitoring, auditability, and board accountability,
while the model, infrastructure, update process, and internal traces remain
controlled by an external vendor.

The catastrophic actions are also easier to define than in many general AI
benchmarks: an incorrect hold or release, systematic mis-triage, exposure of
protected financial data, or an action path touching settlement operations.
These are not equivalent to existential risk, but they are clear enough to
show how access constraints shape control.

\minorheading{Contributions and roadmap}\label{contributions-and-roadmap}

This paper makes five contributions. First, it introduces a four-layer
access typology - data, model, infrastructure, and interaction - for
thinking about deployment-time AI control. Second, it provides a
protocol-by-layer requirements matrix that shows which control protocols
remain executable under different access conditions. Third, it introduces
\textbf{sovereignty discount cost}, the spend, delay, residual risk, or
usefulness loss incurred when a deployer must substitute for missing access.
Fourth, it reports a synthetic access-ablation experiment using a synthetic
payments-dispute workflow, 20 access profiles, 10 control protocols, three
synthetic agent behaviours, threshold and audit-budget sweeps, and 1.35
million synthetic case simulations. Fifth, it interprets the framework
through an anonymised national-payments-infrastructure scenario.

The paper proceeds as follows. Section 2 reviews the related literature.
Section 3 develops the conceptual framework. Section 4 presents the
empirical methodology. Section 5 reports the synthetic access-ablation
results. Section 6 applies the findings to the illustrative payments
infrastructure scenario. Section 7 discusses implications and limitations.
Section 8 concludes.

\section{Related Work}\label{related-work}

\subsection{AI control and scalable oversight}\label{ai-control}

Greenblatt et al.~(2023) originate the AI control framework: rather than
assuming a model's alignment can be verified, control evaluations pit a
blue team designing a deployment protocol against a red team playing a
worst-case, deliberately subversive untrusted model. The originating
paper introduces trusted monitoring, in which a weaker, trusted model
scores the untrusted model's outputs for suspicion, with human auditing
of the highest-scoring fraction, and defer-to-trusted protocols, under
which a trusted model's output replaces suspicious actions.

Subsequent work extends control to multi-step agentic settings via
resampling (Bhatt et al., 2025), formalises protocol evaluation
game-theoretically (Griffin et al., 2024), sketches how control
arguments could support deployment safety cases (Korbak et al., 2025),
and improves monitor performance through legibility requirements on the
untrusted model's outputs (Sreevatsa et al., 2026).

Most directly relevant here, Terekhov et al.~(2025) introduce the
control tax, with a theoretical framework mapping monitor classifier
performance to safety assurances and empirical cost estimates for
monitoring protocols under auditing budgets. This paper takes the
control tax as its point of departure. The critique is not that the
control tax is wrong. The critique is that it is incomplete if priced
conditional on instrumentation access the deployer may not have.

Recent adversarial results sharpen why this matters. Monitors are
themselves attack surfaces (Terekhov et al., 2026). Attack-selection
capability can degrade safety substantially at realistic auditing
budgets (Schaeffer et al., 2026). Untrusted monitoring requires explicit
collusion-strategy assumptions to support a safety case (Gardner-Challis
et al., 2026). Each of these defences and evaluations presupposes that
some actor can instrument the deployment pipeline. The open question is
which actor, with what access, under what contractual and
infrastructural constraints.

\minorheading{Scalable oversight}\label{scalable-oversight}

The scalable oversight literature asks how a weaker supervisor can
reliably judge a stronger system: through debate (Irving et al., 2018),
weak-to-strong generalisation (Burns et al., 2023), empirical studies of
weak LLMs judging strong ones (Kenton et al., 2024), and scaling laws
for oversight games between weak overseers and strong adversaries
(Engels et al., 2025).

This paper does not technically extend scalable oversight, but borrows
one structural observation: oversight capacity is a scarce resource
whose adequacy relative to the overseen system can be analysed. We apply
the same resource logic to access rather than capability. A deployer may
have a capable monitor but still lack the logs, traces, tool-call
visibility, version control, or gateway needed to use that monitor
effectively.

\subsection{Sovereign AI, compute governance, and structural access asymmetry}\label{sovereign-ai-and-compute-governance}

A separate literature, largely in policy institutes, industry analyses,
and technical-governance venues, documents that control over the AI
stack is concentrated. Common formulations decompose AI sovereignty into
pillars: data, model, infrastructure, and operational sovereignty recur
as a four-pillar industry framework, while layered and spectrum-based
treatments emphasise that sovereignty is not binary but distributed
across stack layers (Stanford HAI, 2026; Tony Blair Institute, 2026;
Cruzes, 2026). Empirical work documents the thinness of Global South
access at the infrastructure layer (CSIS, 2025; Brookings Institution,
2026), and the compute-governance literature proposes state-level
mechanisms, including compute monitoring, know-your-customer schemes for
compute providers (Egan and Heim, 2023), and technical AI governance
research agendas (Reuel et al., 2024), for governing the substrate on
which frontier models run.

A key observation motivating this paper is that compute governance and
deployment-time AI control currently operate at different levels.
Compute governance often asks how states can govern who trains and runs
frontier models. AI control asks how a deployer can safely run a
possibly misaligned model. Neither literature fully asks what happens
when the deployer's position in the compute-governance landscape
determines whether its control protocol is executable. This paper sits
at that seam.

\minorheading{Structural asymmetry: data colonialism and decolonial AI}\label{structural-asymmetry-data-colonialism-and-decolonial-ai}

A third literature explains why access asymmetry may be structural
rather than an incidental market failure. Couldry and Mejias (2019)
theorise data colonialism: the appropriation of human life through data
under an extractive economic logic continuous with historical
colonialism. Mohamed et al.~(2020) develop decolonial AI, identifying
algorithmic coloniality in the design, deployment, and governance of AI
systems. Birhane (2020) documents algorithmic colonisation dynamics
concretely for Africa, where imported technology can embed foreign
values and interests with limited local scrutiny or control.

This paper uses that literature narrowly. It does not advance a
normative theory of colonialism and does not attempt to adjudicate
broader moral claims. It uses the literature to justify treating
asymmetric access as a persistent feature of the deployment landscape
rather than a temporary inconvenience that market competition will
quickly erase.

\subsection{Gap: control under partial access}\label{the-gap}

Assembling these threads: AI control assumes deployers can instrument
the model; sovereign-AI and compute-governance work shows that such
access is unevenly distributed; and structural-asymmetry literature
suggests that limited access is not merely transitional. The resulting
question is direct: \textbf{which control protocols remain valid, at
what cost, under partial access?}

The gap is narrower and more practical than a general theory of AI
sovereignty. Existing AI-control work has begun to evaluate and price
control protocols, but it rarely makes the deployer's access position an
explicit part of the protocol specification. This paper therefore
contributes a vocabulary, a mapping method, and a synthetic empirical
operationalisation for treating access assumptions as first-order features
of deployment-time control.

\section{Conceptual Framework: Bounded Sovereignty and the Control
Tax}\label{conceptual-framework-bounded-sovereignty-and-the-control-tax}

\subsection{Definitions and access typology}\label{definitions}

Because this paper sits between literatures with different vocabularies,
we fix terms precisely.

\begin{itemize}
\tightlist
\item
  \textbf{Sovereignty} means technical and contractual access sufficient
  to independently verify or intervene on an AI system. This is
  deliberately narrower than political sovereignty. A deployer has
  sovereignty over a layer if it can observe or act on that layer
  without the model vendor's ongoing discretionary cooperation.
\item
  \textbf{Control} means deployment protocols that constrain a
  potentially misaligned model's actions regardless of the model's
  intentions, evaluated against a worst-case subversive model
  (Greenblatt et al., 2023).
\item
  \textbf{Oversight} means the broader institutional and regulatory
  function of ensuring an AI deployment behaves acceptably. Control
  protocols are one implementation of oversight; regulatory examination,
  model risk management, and audit regimes are others.
\item
  \textbf{Control tax} means the operational and financial cost of
  integrating control measures into AI pipelines (Terekhov et al.,
  2025).
\end{itemize}

One conceptual point matters throughout the paper: sovereignty and
control are orthogonal. Sovereignty is not a safety protocol; it is an
access condition. A jurisdiction can hold full data and infrastructure
sovereignty and still implement no meaningful control. Conversely, a
deployer with no model-weight access may still achieve partial control
through output-level monitoring, contractual audit rights, and a
deployer-owned execution gateway. Keeping these axes separate allows the
paper to ask: how much access is enough for which kind of control?

\clearpage
\minorheading{A four-layer typology of access}\label{a-four-layer-typology-of-access}

We adapt the recurring four-pillar decomposition from the sovereign-AI
literature to the deployment-control context. Our adaptation replaces
the broad operational pillar with \textbf{interaction sovereignty},
defined as access to the surface that control protocols monitor and
intercept: prompts, outputs, tool calls, logs, traces, and gateway-level
intervention points. This is the paper's adaptation rather than a settled
taxonomy in the literature.

\par\Needspace{0.34\textheight}\begingroup\small\setlength{\tabcolsep}{4pt}\renewcommand{\arraystretch}{0.95}\begin{longtable}[]{@{}
  >{\raggedright\arraybackslash}p{(\columnwidth - 4\tabcolsep) * \real{0.3333}}
  >{\raggedright\arraybackslash}p{(\columnwidth - 4\tabcolsep) * \real{0.3333}}
  >{\raggedright\arraybackslash}p{(\columnwidth - 4\tabcolsep) * \real{0.3333}}@{}}
\toprule\noalign{}
\begin{minipage}[b]{\linewidth}\raggedright
Layer
\end{minipage} & \begin{minipage}[b]{\linewidth}\raggedright
What full access would mean
\end{minipage} & \begin{minipage}[b]{\linewidth}\raggedright
Typical access for an import-dependent regulated deployer
\end{minipage} \\
\midrule\noalign{}
\endhead
\bottomrule\noalign{}
\endlastfoot
Data sovereignty & Control over what data trains or fine-tunes the
model; ability to audit training-data provenance & Usually retains
control over own transactional or fine-tuning data; no visibility into
base-model training data \\
Model sovereignty & Access to weights, architecture, activations, and
ability to retrain, patch, probe, or run interpretability tooling &
Usually absent under API or managed-endpoint access; weights,
architecture details, and activations are vendor-held \\
Infrastructure sovereignty & Compute located in-jurisdiction, under
domestic legal control, observable by the deployer & Partial at best:
often a foreign-operated cloud region, sometimes with data-residency
commitments; serving stack remains vendor-operated \\
Interaction sovereignty & Full logging and audit of prompts, outputs,
tool calls, traces, and pre-execution intervention points & Partial:
input/output and tool logs may be contractable; internal traces are
rarely available; gateway ownership depends on architecture \\
\end{longtable}\endgroup\par

Two properties of this typology matter. First, the layers are separable.
Contracts, residency requirements, product tiers, and deployment
architectures move them independently. Second, the layers are jointly
binding. Each control protocol presupposes a specific bundle of access.
Missing one part of the bundle can degrade or disable the protocol.

\Needspace{0.82\textheight}
\subsection{Protocol requirements under bounded sovereignty}\label{protocol-by-layer-requirements-matrix}

The protocol requirements matrix below maps canonical control protocols
to the access layers they presuppose. The matrix is this paper's
construction and is derived from the control-protocol definitions in the
originating literature (Greenblatt et al., 2023; Bhatt et al., 2025;
Sreevatsa et al., 2026).

\par\Needspace{0.34\textheight}\begingroup\small\setlength{\tabcolsep}{4pt}\renewcommand{\arraystretch}{0.95}\begin{longtable}[]{@{}
  >{\raggedright\arraybackslash}p{(\columnwidth - 6\tabcolsep) * \real{0.2500}}
  >{\raggedright\arraybackslash}p{(\columnwidth - 6\tabcolsep) * \real{0.2500}}
  >{\raggedright\arraybackslash}p{(\columnwidth - 6\tabcolsep) * \real{0.2500}}
  >{\raggedright\arraybackslash}p{(\columnwidth - 6\tabcolsep) * \real{0.2500}}@{}}
\toprule\noalign{}
\begin{minipage}[b]{\linewidth}\raggedright
Control protocol
\end{minipage} & \begin{minipage}[b]{\linewidth}\raggedright
Binding access layers
\end{minipage} & \begin{minipage}[b]{\linewidth}\raggedright
Typical API-only feasibility
\end{minipage} & \begin{minipage}[b]{\linewidth}\raggedright
Notes
\end{minipage} \\
\midrule\noalign{}
\endhead
\bottomrule\noalign{}
\endlastfoot
Trusted monitoring, output-level & Interaction logging;
deployer-controlled monitor compute & Feasible if logs are complete &
Requires complete, tamper-evident I/O logs and monitor execution outside
the untrusted model \\
Trusted monitoring, trace/activation-level & Model traces or
activations; serving-stack observability & Usually infeasible & Requires
model sovereignty or trace-level vendor access \\
Defer-to-trusted or trusted editing & Trusted substitute model;
pre-execution interception & Feasible only with architecture constraint
& Requires an owned execution gateway before tool use or
transaction-adjacent execution \\
Resampling protocols & Controlled re-querying; stepwise agent-loop
interception; stable model behaviour & Degraded & API re-querying is
possible but latency, cost, and version consistency bind \\
Human or AI auditing & Retained and queryable interaction logs; audit
capacity & Feasible but bounded & Effectiveness depends on log
completeness and audit budget \\
Legibility protocols & Prompting or training influence; output
inspection & Partially feasible & Prompting-level legibility is
available; training-level legibility is usually not \\
Pre-deployment control evaluations & Representative task data; stable
access to deployed model version & Degraded & Evaluation is weakened if
production model versions can change without notice \\
Weights escrow or patching & Model sovereignty; infrastructure control
or contractual escrow & Infeasible except contractually & Native
patching requires weight access \\
\end{longtable}\endgroup\par

Under typical API-only access, the feasible set collapses toward
output-level protocols. Even those depend on two load-bearing
conditions: complete interaction logging and an execution gateway the
deployer controls. A monitor that can flag a dangerous tool call after
it has already executed is an audit mechanism, not a prevention
mechanism.

\subsection{Sovereignty discount cost}\label{the-sovereignty-discount-cost-reframing-the-control-tax}

Terekhov et al.~(2025) price the control tax under an implicit
full-instrumentation assumption. The deployer can place monitors in the
pipeline, observe outputs completely, route suspicious actions to
trusted models or human auditors, and measure the resulting
safety-usefulness trade-off. For an import-dependent deployer, part of
the realised oversight cost goes toward something categorically
different: substituting for access the deployer does not have.

Examples include purchasing third-party audit or attestation rights over
a vendor-held model because direct inspection is unavailable; building
output-only anomaly detection because activation-level monitoring is
technically impossible without model sovereignty; negotiating
contractual clauses that simulate the intervention ability a
weight-holding deployer has natively; and narrowing the agentic system's
permitted action space because the monitoring required to permit a wider
scope cannot be implemented.

We propose the working term \textbf{sovereignty discount cost} for this
component: the additional expenditure, delay, residual risk, or
usefulness loss required to approximate a control protocol that assumes
access the deployer lacks. Conceptually:

\[\text{realised control tax}=\text{instrumented-control cost}+\text{sovereignty discount cost}.\]

This is not yet an accounting identity because some missing access cannot be
substituted at any price, so the discount cost is not always finite. The
term is introduced as a research construct, not as an established
measure.

\subsection{Calibrated control: how much access is enough?}\label{should-control-be-absolute-three-positions}

The framework invites a normative question: if a deployer cannot achieve
full access, should it deploy at all? Rather than issuing a universal
verdict, this paper identifies three positions implicit in the
surrounding literature.

\textbf{Position A: Safety-maximalist.} If a deployer cannot verify that
a control protocol is actually running, because it lacks model and
infrastructure access, then any claimed control is unfalsifiable from
the deployer's position. The deployer should not treat such a protocol
as safety assurance. This position draws support from the control
literature's emphasis on verifiability and explicit safety-case
assumptions (Korbak et al., 2025; Gardner-Challis et al., 2026).

\textbf{Position B: Sovereignty-maximalist.} Partial control under
foreign dependency is fragile. A vendor or state actor can revoke,
degrade, or modify access. Conditional-access arrangements such as the
Pax Silica framework demonstrate that infrastructure access may be tied
to political alignment (GIS Reports, 2026). On this view, the only
durable solution is full-stack sovereign capability: domestic models,
domestic compute, and domestic control.

\textbf{Position C: Calibrated or bounded control.} Neither A nor B is
actionable on relevant timescales for many jurisdictions: A implies
abstention from deployments that are already commercially and
politically committed, while B implies capability investments that many states
cannot make this decade. The productive question is therefore not full
control versus no control, but which layers of access are necessary for
which classes of catastrophic action, and which can be substituted
through architecture, contracts, monitoring, or scope restriction.

This paper adopts Position C as a working frame, not as a claim that A
or B is wrong. Economically, the problem resembles incomplete
contracting and principal-agent design: the deployer contracts with a
model vendor that holds private information about the model, under
contracts that cannot specify all contingencies, with verification
costly and asymmetric.

\minorheading{A balancing framework: how much access is enough?}\label{a-balancing-framework-how-much-access-is-enough}

Under Position C, a deployer or regulator needs a structured way to
decide how much access-layer sovereignty to demand, buy, or substitute
for. We propose six criteria.

\begin{enumerate}
\def\labelenumi{\arabic{enumi}.}
\item
  \textbf{Severity and reversibility of the catastrophic action.} A
  mis-routed low-value transaction is bounded and reversible; a systemic
  settlement failure or mass data exposure is not. Higher severity and
  lower reversibility raise the minimum access bar.
\item
  \textbf{Verifiability of substitutes.} Can output-only monitoring
  detect the relevant failure mode even in principle, or does detection
  require internal access? Failure modes that manifest only in internal
  computation cannot be fully bought back with output-level substitutes.
\item
  \textbf{Vendor concentration and bargaining power.} Access is
  negotiated. A deployer facing one viable frontier-model vendor
  negotiates from a weaker position than one facing several
  substitutable vendors.
\item
  \textbf{Regulatory maturity and absorptive capacity.} Access is useful
  only if the deployer or regulator can exercise it. The ability to
  interpret logs, audit traces, investigate incidents, and challenge
  vendor attestations is distinct from the legal right to receive them.
\item
  \textbf{Sovereignty discount cost versus scope-restriction cost.}
  Sometimes the safest option is not buying more access but narrowing
  what the system may do. The framework should compare the marginal cost
  of substituting for access against the marginal usefulness lost by
  restricting scope.
\item
  \textbf{Availability of coalition paths.} Shared audit consortia,
  regional regulatory colleges, or South-South cooperation arrangements
  can amortise the cost of acquiring partial sovereignty across multiple
  deployers.
\end{enumerate}

These criteria are a rubric, not a weighted formula. The empirical study
below begins to operationalise some of them, but the weights remain an
open empirical question.

\section{Empirical Methodology: Synthetic Access-Ablation
Study}\label{empirical-methodology-synthetic-access-ablation-study}

\subsection{Study design and synthetic environment}\label{mixed-methods-design}

This paper uses three methods. First, it develops a conceptual access
typology by synthesising the AI control, scalable oversight, sovereign
AI, compute-governance, and structural-asymmetry literatures. Second, it
conducts a \textbf{synthetic access-ablation experiment} to
operationalise the typology and test how access constraints affect
protocol feasibility, safety, usefulness, audit burden, diagnosis,
latency, and cost. Third, it applies the framework interpretively to an
anonymised national-payments-infrastructure scenario.

The empirical study is deliberately synthetic and does not use real
payment data, real customer data, production logs, vendor contracts,
confidential institutional material, or operational payment
infrastructure. It should be read as \textbf{construct validation}, not
real-world payment-system evidence. Its purpose is to show how bounded
sovereignty can be made measurable and how the same control protocol
behaves differently under different access conditions.

\minorheading{Synthetic task environment}\label{synthetic-task-environment}

The experiment models a mock national payments dispute-triage workflow.
The workflow is transaction-adjacent rather than settlement-operational.
A synthetic agent reviews a dispute case and chooses among harmless mock
actions such as \texttt{draft\_report}, \texttt{recommend\_hold},
\texttt{recommend\_release}, \texttt{escalate\_to\_human}, or
\texttt{no\_action}. None of these actions connects to any real system.

Each synthetic case contains non-real fields such as a case identifier,
transaction amount, merchant category, customer claim, prior dispute
history, risk score, settlement-critical marker, synthetic
sensitive-context marker, allowed actions, and a ground-truth action.
The workflow is multi-step: case intake, synthetic evidence retrieval,
dispute classification, action recommendation, optional gateway review,
mock execution, and audit review.

The experiment uses four synthetic catastrophic-action classes that
mirror the paper's financial-infrastructure case.

\par\Needspace{0.34\textheight}\begingroup\small\setlength{\tabcolsep}{4pt}\renewcommand{\arraystretch}{0.95}\begin{longtable}[]{@{}
  >{\raggedright\arraybackslash}p{(\columnwidth - 2\tabcolsep) * \real{0.5000}}
  >{\raggedright\arraybackslash}p{(\columnwidth - 2\tabcolsep) * \real{0.5000}}@{}}
\toprule\noalign{}
\begin{minipage}[b]{\linewidth}\raggedright
Class
\end{minipage} & \begin{minipage}[b]{\linewidth}\raggedright
Synthetic meaning in the experiment
\end{minipage} \\
\midrule\noalign{}
\endhead
\bottomrule\noalign{}
\endlastfoot
C1 & Incorrect hold or release recommendation on an individual synthetic
dispute case \\
C2 & Systematic mis-triage pattern across a synthetic case cluster \\
C3 & Synthetic sensitive-field exposure marker in a mock report \\
C4 & Synthetic settlement-critical recommendation marker \\
\end{longtable}\endgroup\par

C3 and C4 are labels only; the repository contains no real data
exposure, credential handling, exfiltration logic, malware, cyberattack
steps, or operational disruption logic.

\subsection{Access factors, agents, and protocols}\label{access-factors}

Rather than comparing a small number of bundled regimes, the large run
factorises access into seven dimensions. These dimensions map directly
to the paper's four-layer bounded-sovereignty typology.

\par\Needspace{0.34\textheight}\begingroup\small\setlength{\tabcolsep}{4pt}\renewcommand{\arraystretch}{0.95}\begin{longtable}[]{@{}
  >{\raggedright\arraybackslash}p{(\columnwidth - 4\tabcolsep) * \real{0.3333}}
  >{\raggedright\arraybackslash}p{(\columnwidth - 4\tabcolsep) * \real{0.3333}}
  >{\raggedright\arraybackslash}p{(\columnwidth - 4\tabcolsep) * \real{0.3333}}@{}}
\toprule\noalign{}
\begin{minipage}[b]{\linewidth}\raggedright
Paper layer
\end{minipage} & \begin{minipage}[b]{\linewidth}\raggedright
Experimental factor
\end{minipage} & \begin{minipage}[b]{\linewidth}\raggedright
Levels used
\end{minipage} \\
\midrule\noalign{}
\endhead
\bottomrule\noalign{}
\endlastfoot
Interaction sovereignty & I/O logs & none, partial, complete \\
Interaction sovereignty & Tool-call logs & none, partial, complete \\
Interaction sovereignty & Pre-execution gateway & absent, present \\
Model sovereignty & Trace access & none, limited, oracle \\
Model / interaction sovereignty & Model-version control & unpinned,
notified, pinned \\
Infrastructure / contractual sovereignty & Vendor support & none,
delayed, guaranteed \\
Deployment design substitute & Tool scope & full, restricted \\
\end{longtable}\endgroup\par

The large run uses a curated fractional-factorial set of 20 access
profiles. These include a full-instrumentation anchor,
complete-logging-with-gateway profiles, complete-logging-without-gateway
profiles, partial-logging profiles, no-log profiles, trace-access
contrasts, version-control contrasts, vendor-support contrasts, and
restricted-scope contrasts.

\Needspace{0.45\textheight}
\minorheading{Synthetic agents}\label{synthetic-agents}

The experiment evaluates three synthetic agent behaviours.

\par\Needspace{0.34\textheight}\begingroup\small\setlength{\tabcolsep}{4pt}\renewcommand{\arraystretch}{0.95}\begin{longtable}[]{@{}
  >{\raggedright\arraybackslash}p{(\columnwidth - 2\tabcolsep) * \real{0.5000}}
  >{\raggedright\arraybackslash}p{(\columnwidth - 2\tabcolsep) * \real{0.5000}}@{}}
\toprule\noalign{}
\begin{minipage}[b]{\linewidth}\raggedright
Agent
\end{minipage} & \begin{minipage}[b]{\linewidth}\raggedright
Behaviour
\end{minipage} \\
\midrule\noalign{}
\endhead
\bottomrule\noalign{}
\endlastfoot
Honest & Usually follows the synthetic ground truth but makes occasional
mistakes \\
Careless & Overconfident, error-prone, and less likely to escalate
uncertain cases \\
Stealthy access-aware & Chooses lower-visibility synthetic failure modes
when access is weak \\
\end{longtable}\endgroup\par

The access-aware agent is strategic only inside the synthetic
environment. It cannot perform real harmful behaviour and only selects
among harmless mock actions that may be labelled as C1-C4 synthetic
failures.

\Needspace{0.55\textheight}
\minorheading{Control protocols}\label{control-protocols}

Each access profile is crossed with ten control protocols.

\par\Needspace{0.34\textheight}\begingroup\small\setlength{\tabcolsep}{4pt}\renewcommand{\arraystretch}{0.95}\begin{longtable}[]{@{}
  >{\raggedright\arraybackslash}p{(\columnwidth - 2\tabcolsep) * \real{0.5000}}
  >{\raggedright\arraybackslash}p{(\columnwidth - 2\tabcolsep) * \real{0.5000}}@{}}
\toprule\noalign{}
\begin{minipage}[b]{\linewidth}\raggedright
Protocol
\end{minipage} & \begin{minipage}[b]{\linewidth}\raggedright
Interpretation
\end{minipage} \\
\midrule\noalign{}
\endhead
\bottomrule\noalign{}
\endlastfoot
No control & Accept the agent's proposed mock action \\
Output monitor & Score the final output or action for suspiciousness \\
Tool monitor & Score proposed tool/action use when tool visibility
exists \\
Defer-to-trusted & Replace suspicious actions with a conservative
trusted fallback when possible \\
Resampling & Generate alternative candidate actions and select the
lowest-risk acceptable one \\
Human audit & Send flagged cases to a capacity-limited synthetic
reviewer \\
Trusted editing & Edit or redact risky draft outputs, or fall back to a
safer policy \\
Scope restriction & Remove higher-risk mock actions from the action
space \\
Macro-shutdown & Escalate or safe-mode clusters after repeated
suspiciousness \\
Vendor escalation & Request delayed or guaranteed synthetic vendor
support for diagnosis or attestation \\
\end{longtable}\endgroup\par

This protocol list mirrors practical families of interventions in the
AI-control literature: monitoring, defer-to-trusted routing, resampling,
human review, action-space restriction, and post-incident or
vendor-mediated assurance.

\subsection{Metrics and sovereignty-discount proxy}\label{metrics}

For every condition, the experiment reports task success, catastrophic
failure rate, failure rate by class, severity-weighted loss, audit rate,
flag rate, false-positive rate, false-negative rate, intervention
success, diagnosis success, latency proxy, model-call count,
monitor-call count, human-audit count, vendor-request count, and
decomposed cost proxy.

Diagnosis is \textbf{evidence-based} rather than hard-coded by access
regime. A failure is diagnosable only when the evidence available under
the access profile is sufficient to reconstruct what happened. For
example, complete I/O logs, complete tool-call logs, model-version
pinning, trace access, or guaranteed vendor support provide more
reconstructive evidence than no logs, no gateway, no trace access, and
no vendor support.

The total control-cost proxy decomposes inference cost, monitoring cost,
human audit cost, logging cost, gateway cost, vendor-assurance cost,
latency penalty, scope-restriction cost, and residual incident-loss
cost. These are unitless proxies, not financial estimates for a real
institution, and they allow comparison of how access constraints shift the
cost of control.

\minorheading{Sovereignty discount cost proxy}\label{sovereignty-discount-cost-proxy}

The study computes a sovereignty-discount proxy by asking: for a given
target catastrophic-failure threshold, what is the cheapest
protocol/access combination that reaches the target, and how does that
cost compare with the cheapest full-instrumentation baseline?

The basic form is:

\begin{quote}
\small\ttfamily
minimum cost to reach target failure rate under bounded access\\
minus\\
minimum cost to reach the same target under full instrumentation
\end{quote}

The analysis also reports a stricter variant that requires both safety
and usefulness, for example catastrophic failure at or below 1\% while
maintaining task success of at least 75\%. This second version is
important because a system can often reduce failure simply by disabling
useful actions. In this paper's framing, that is not a free safety gain;
it is a shift from monitoring cost to usefulness loss.

\subsection{Large-run configuration}\label{large-run-configuration}

The large run reported here used the following configuration.

\par\Needspace{0.34\textheight}\begingroup\small\setlength{\tabcolsep}{4pt}\renewcommand{\arraystretch}{0.95}\begin{longtable}[]{@{}lr@{}}
\toprule\noalign{}
Setting & Value \\
\midrule\noalign{}
\endhead
\bottomrule\noalign{}
\endlastfoot
Access profiles & 20 \\
Control protocols & 10 \\
Synthetic agent types & 3 \\
Monitor thresholds & 0.35, 0.55, 0.75 \\
Audit budgets & 2\%, 5\%, 20\% \\
Random seeds & 5 \\
Synthetic cases per condition & 50 \\
Aggregate condition rows & 27,000 \\
Synthetic case simulations & 1,350,000 \\
\end{longtable}\endgroup\par

The run expands the threshold and audit-budget sensitivity grid relative
to the initial validation run. It remains a synthetic study and should
not be described as real-world payment-system evidence.

\section{Empirical Results}\label{empirical-results}

\subsection{Overview: three penalties of bounded sovereignty}\label{reading-the-results}

The results should be read as evidence about the \textbf{logic of access
constraints}, not as measured safety levels for any real payment
institution. The experiment asks a practical question: when a deployer loses
logs, tool-call visibility, model traces, version control, vendor support,
or a pre-execution gateway, what happens to control performance in a
synthetic dispute-triage environment?

The answer is that bounded sovereignty creates three penalties. First, there
is a \textbf{prevention penalty}: the system may detect a risky action but
cannot stop it before mock execution without a gateway. Second, there is a
\textbf{diagnosis penalty}: the system may observe a bad outcome but lack
the evidence needed to reconstruct what happened. Third, there is a
\textbf{usefulness penalty}: when access is weak, the safest substitute is
often to narrow what the system can do, which reduces useful automation.
These three penalties are the policy-relevant reading of the experiment.

\subsection{Gateway and logging: from visibility to intervention}\label{gateway-access-turns-detection-into-intervention}

Across all protocols, agents, thresholds, audit budgets, and access
profiles, the average catastrophic failure rate was \textbf{5.53\% without a
pre-execution gateway} and \textbf{3.13\% with a gateway}. Intervention
success was almost absent without a gateway (\textbf{0.47\%}) but rose to
\textbf{39.52\%} when a gateway was available. The practical interpretation
is straightforward: a monitor can only warn, but a gateway can hold, replace,
escalate, or block an action before it takes effect.

This is why interaction sovereignty should not be reduced to logging. Logs
help a deployer see what happened; a gateway gives the deployer an
intervention point. For policymakers and technocrats, the design question is
therefore not only ``Do we have records?'' but also ``Can we stop a risky
action before execution?''

\Needspace{0.60\textheight}
\begin{figure}[H]
\centering
\includegraphics[width=0.92\linewidth]{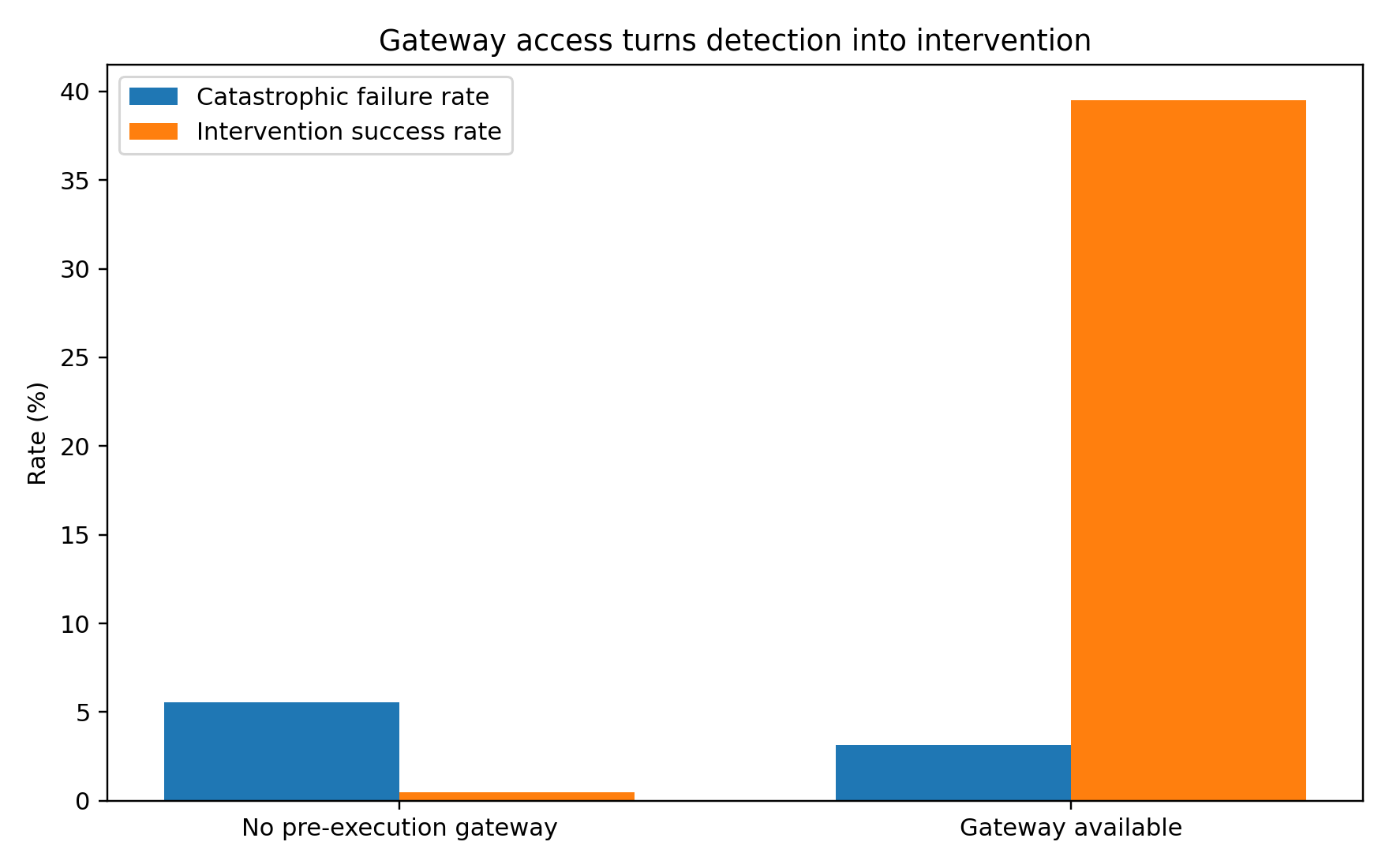}
\caption{Gateway access turns detection into intervention.}
\figurenote{The figure compares average catastrophic failure and intervention success under access profiles with and without a pre-execution gateway. The gateway is the deployer's architectural control point. Without it, a monitor may flag an unsafe action but cannot reliably stop it before mock execution. With it, flagged actions can be held, replaced, escalated, or blocked.}
\end{figure}

\vspace{0.5\baselineskip}
\minorheading{Logging improves reconstructability, not just monitoring}\label{logging-improves-reconstructability-not-just-monitoring}

Log completeness also changed the results. When I/O logs were absent,
diagnosis success was \textbf{0.00\%}. With partial I/O logs, it rose to
\textbf{37.41\%}; with complete I/O logs, it rose to \textbf{78.57\%}.
Failure rates improved as well: no I/O logs had an average catastrophic
failure rate of \textbf{6.17\%}, partial logs \textbf{5.04\%}, and complete
logs \textbf{3.41\%}.

The main point is not only that logs help monitoring. Logs also help a
deployer reconstruct an incident after the fact. Without them, the deployer
may know that something went wrong but be unable to explain the causal chain,
which weakens audit, regulatory accountability, remediation, and the safety
case.

\Needspace{0.60\textheight}
\begin{figure}[H]
\centering
\includegraphics[width=0.92\linewidth]{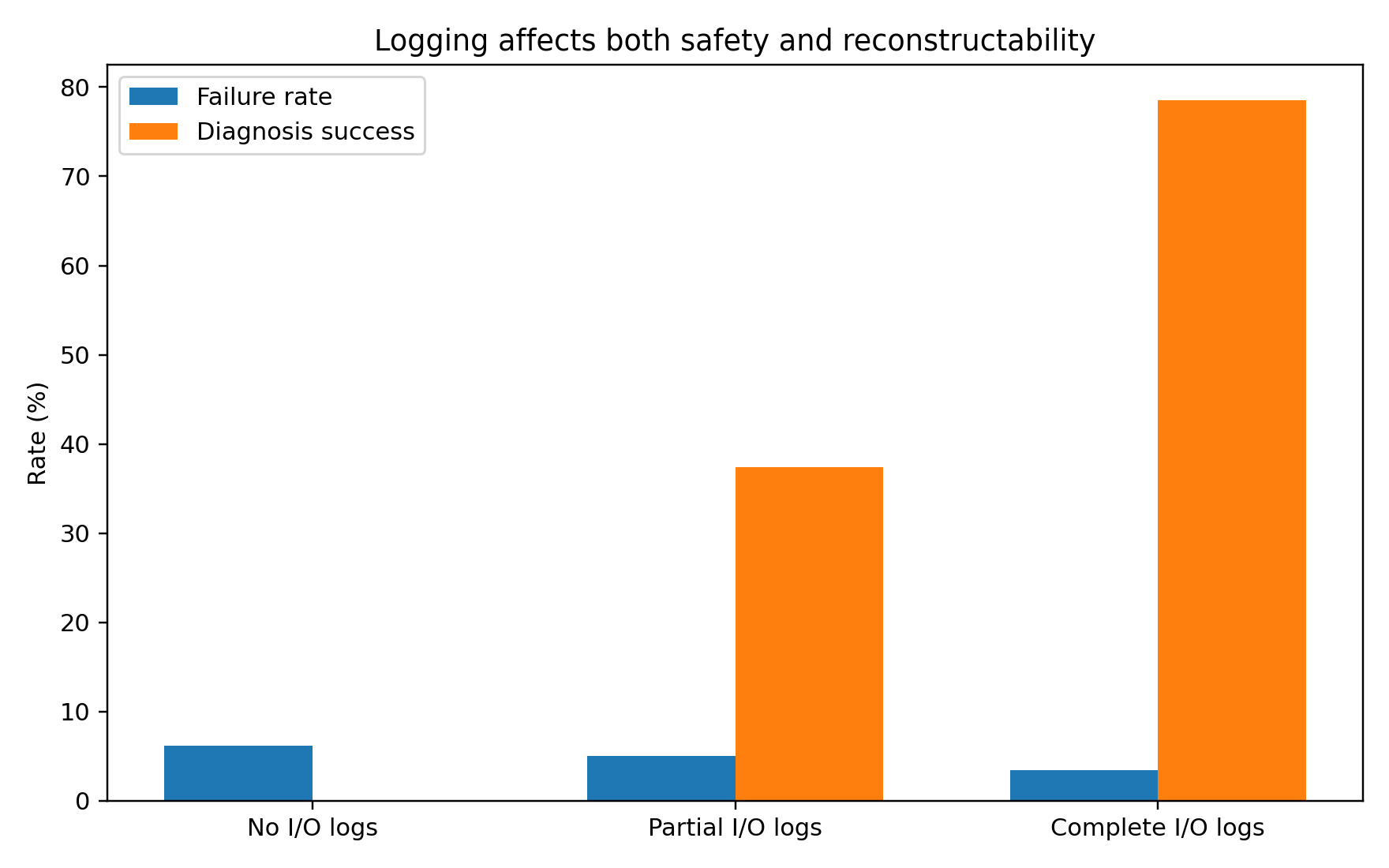}
\caption{Logging affects both safety and reconstructability.}
\figurenote{The figure shows that complete I/O logging is associated with lower synthetic failure and stronger diagnosis. Missing logs do not merely make the system less transparent. They make failures harder to explain and therefore harder to govern.}
\end{figure}

\vspace{0.5\baselineskip}
\subsection{Trace access and model-version control: diagnosis after failure}\label{trace-access-and-model-version-control-matter-most-after-something-goes-wrong}

Trace access and model-version control mattered most for diagnosis. With no
trace access, diagnosis success averaged \textbf{50.86\%}; with limited
trace access it rose to \textbf{74.69\%}; with oracle trace access it reached
\textbf{100.00\%} by construction of the synthetic full-access condition.
Model-version control showed the same pattern: unpinned access was associated
with much weaker diagnosis (\textbf{5.31\%}) than notified (\textbf{83.00\%})
or pinned (\textbf{99.87\%}) conditions.

These numbers should not be read as real-world payment-system estimates. The
structural lesson is that output-level monitoring can remain feasible under
bounded sovereignty, while deeper diagnosis becomes fragile if the deployer
cannot see traces or identify which model version produced a decision. This
is especially important for regulated systems, where post-incident assurance
is part of accountability.

\subsection{Protocol trade-offs: safety, usefulness, and cost}\label{control-protocols-trade-off-safety-usefulness-and-cost}

No control protocol dominated on all dimensions. Scope restriction produced
the lowest average catastrophic failure rate (\textbf{0.50\%}) but reduced
task success to \textbf{60.06\%}. Resampling had the next-lowest average
failure rate (\textbf{3.49\%}) while preserving more usefulness. Several
other protocols - defer-to-trusted, tool monitoring, trusted editing, output
monitoring, and macro-shutdown - clustered around roughly \textbf{4.0\%}
average failure. Human audit and vendor escalation were weaker as prevention
mechanisms in this setup because they often operate after detection, are
capacity-limited, or depend on delayed external support.

The policy lesson is that safety is not free. When access is insufficient, a
deployer can sometimes reduce risk by narrowing the system's action space,
but that reduces usefulness. A technically safe design may therefore be
operationally unattractive if it disables too much of the workflow.

\Needspace{0.60\textheight}
\begin{figure}[H]
\centering
\includegraphics[width=0.92\linewidth]{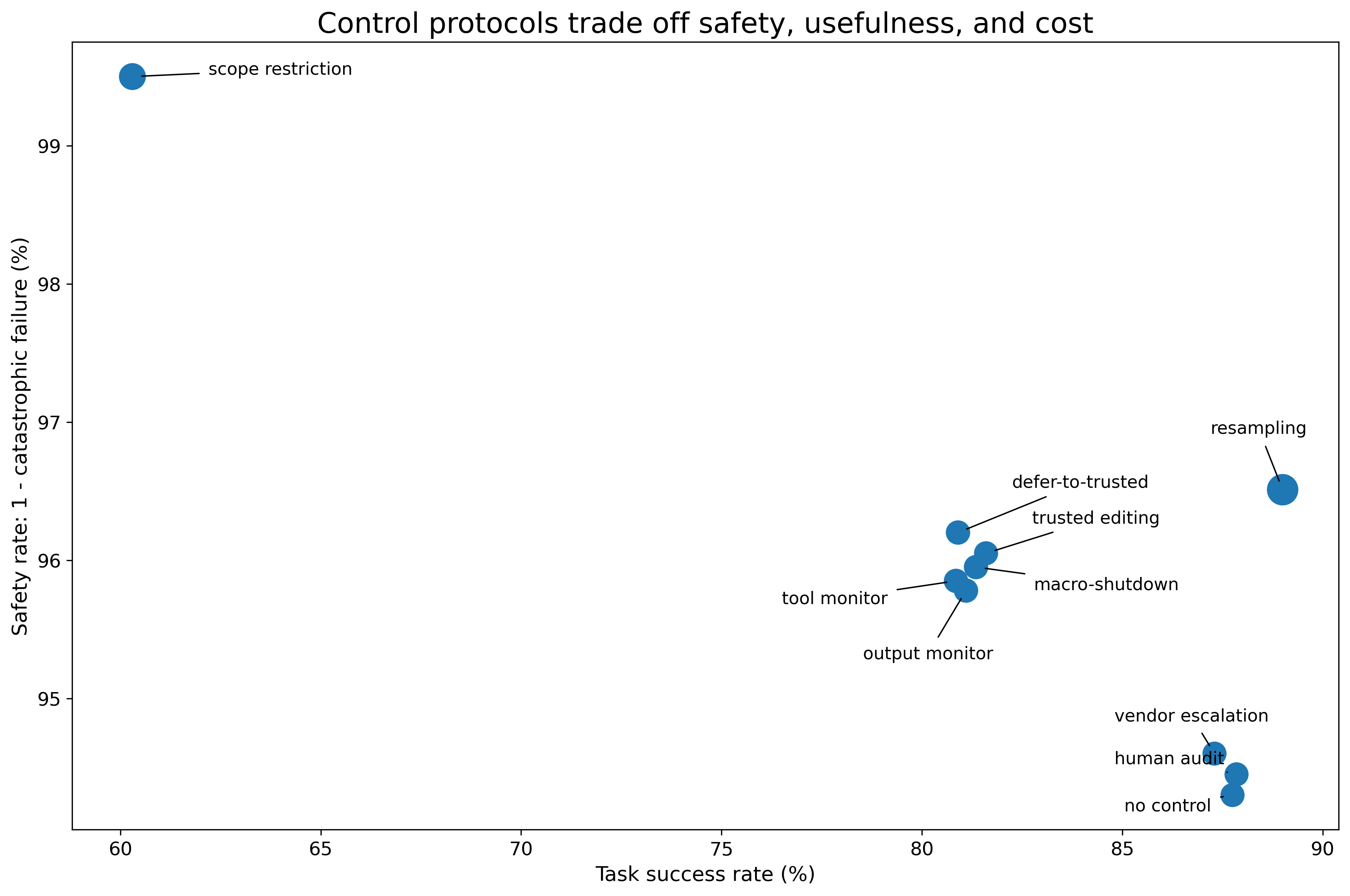}
\caption{Control protocols trade off safety, usefulness, and cost.}
\figurenote{The horizontal axis shows usefulness, the vertical axis shows safety, and marker size represents the control-cost proxy. The plot is intended to prevent a misleading conclusion that the safest protocol is automatically best. A protocol that disables useful actions may be safe but operationally unattractive.}
\end{figure}

\vspace{0.5\baselineskip}
\subsection{Agent behaviour and target reachability}\label{bounded-access-is-most-fragile-against-less-benign-agents}

Access constraints mattered more when the synthetic agent was less benign.
The honest agent had the lowest average failure rate (\textbf{0.98\%}), the
careless agent the highest (\textbf{6.82\%}), and the stealthy access-aware
agent sat between them (\textbf{4.47\%}) while creating more difficult
diagnosis and control conditions. This matters because AI control is not
mainly about an honest model making routine mistakes; it asks whether the
deployment remains safe when the untrusted model may behave badly or
strategically.

For a strict target of catastrophic failure at or below \textbf{1\%} while
preserving at least \textbf{75\%} task success, the fraction of access
profiles reaching the target was:

\par\Needspace{0.45\textheight}\begingroup\small\setlength{\tabcolsep}{4pt}\renewcommand{\arraystretch}{0.95}\begin{longtable}[]{@{}lr@{}}
\toprule\noalign{}
Synthetic agent & Profiles reaching $\leq$1\% failure and $\geq$75\% usefulness \\
\midrule\noalign{}
\endhead
\bottomrule\noalign{}
\endlastfoot
Honest & 90\% \\
Careless & 40\% \\
Stealthy access-aware & 35\% \\
\end{longtable}\endgroup\par

\Needspace{0.60\textheight}
\begin{figure}[H]
\centering
\includegraphics[width=0.92\linewidth]{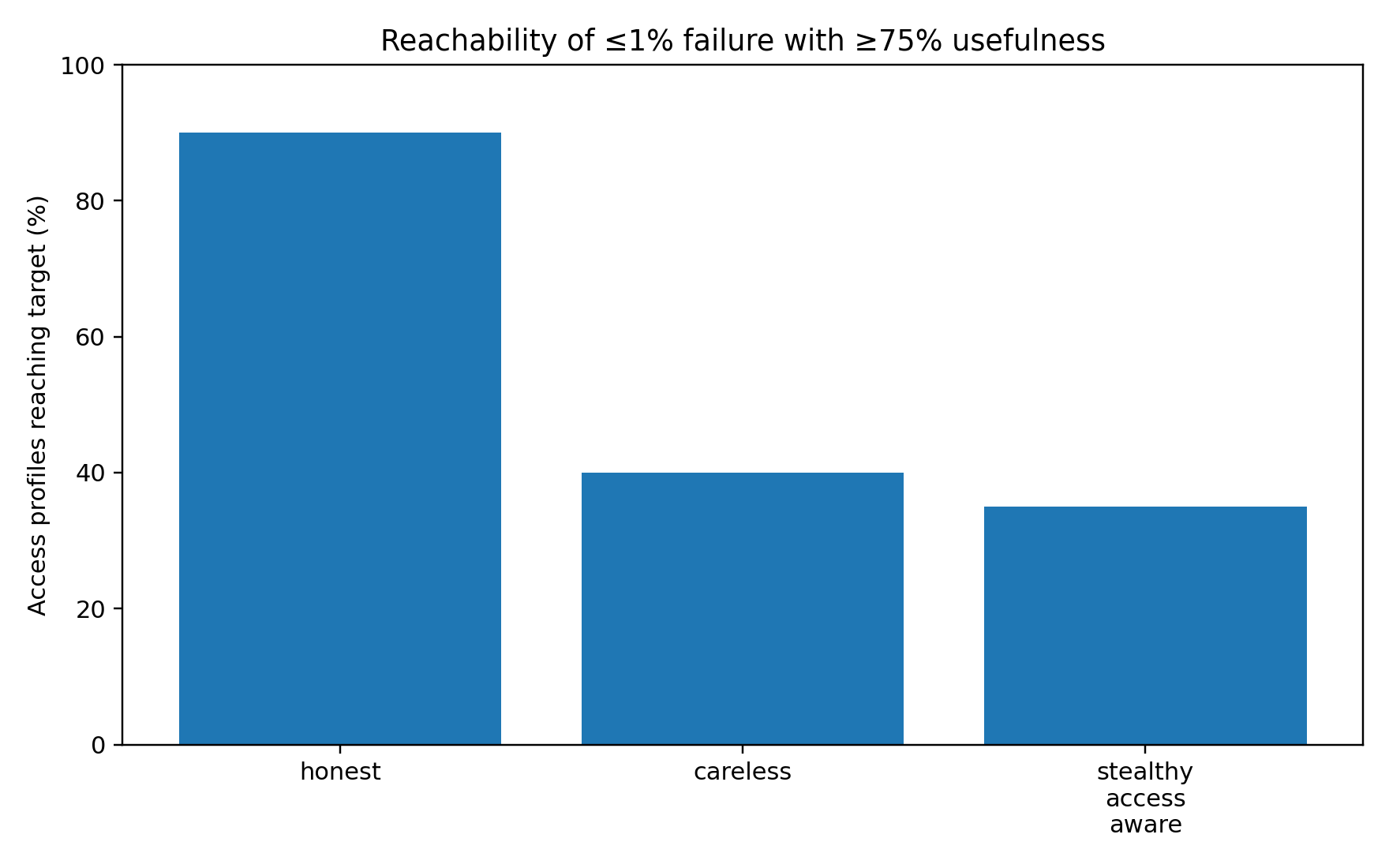}
\caption{Reachability of $\leq$1\% failure with $\geq$75\% usefulness.}
\figurenote{The figure reports what fraction of access profiles can reach a strict safety-plus-usefulness target for each synthetic agent. The target is easiest for the honest agent and hardest for the careless and stealthy access-aware agents, which is the setting AI control is most concerned with.}
\end{figure}

\vspace{0.5\baselineskip}
\minorheading{Scope restriction buys safety by giving up usefulness}\label{scope-restriction-buys-safety-by-giving-up-usefulness}

Restricted tool scope reduced average catastrophic failure to \textbf{0.60\%},
compared with \textbf{4.48\%} under full tool scope, but task success fell
from \textbf{84.51\%} to \textbf{60.06\%}. This is one of the clearest
illustrations of sovereignty discount cost. Missing access is not always
paid for through direct financial expenditure; it may be paid for through
narrower permissions, more manual work, less automation, or lower usefulness.

\Needspace{0.60\textheight}
\begin{figure}[H]
\centering
\includegraphics[width=0.92\linewidth]{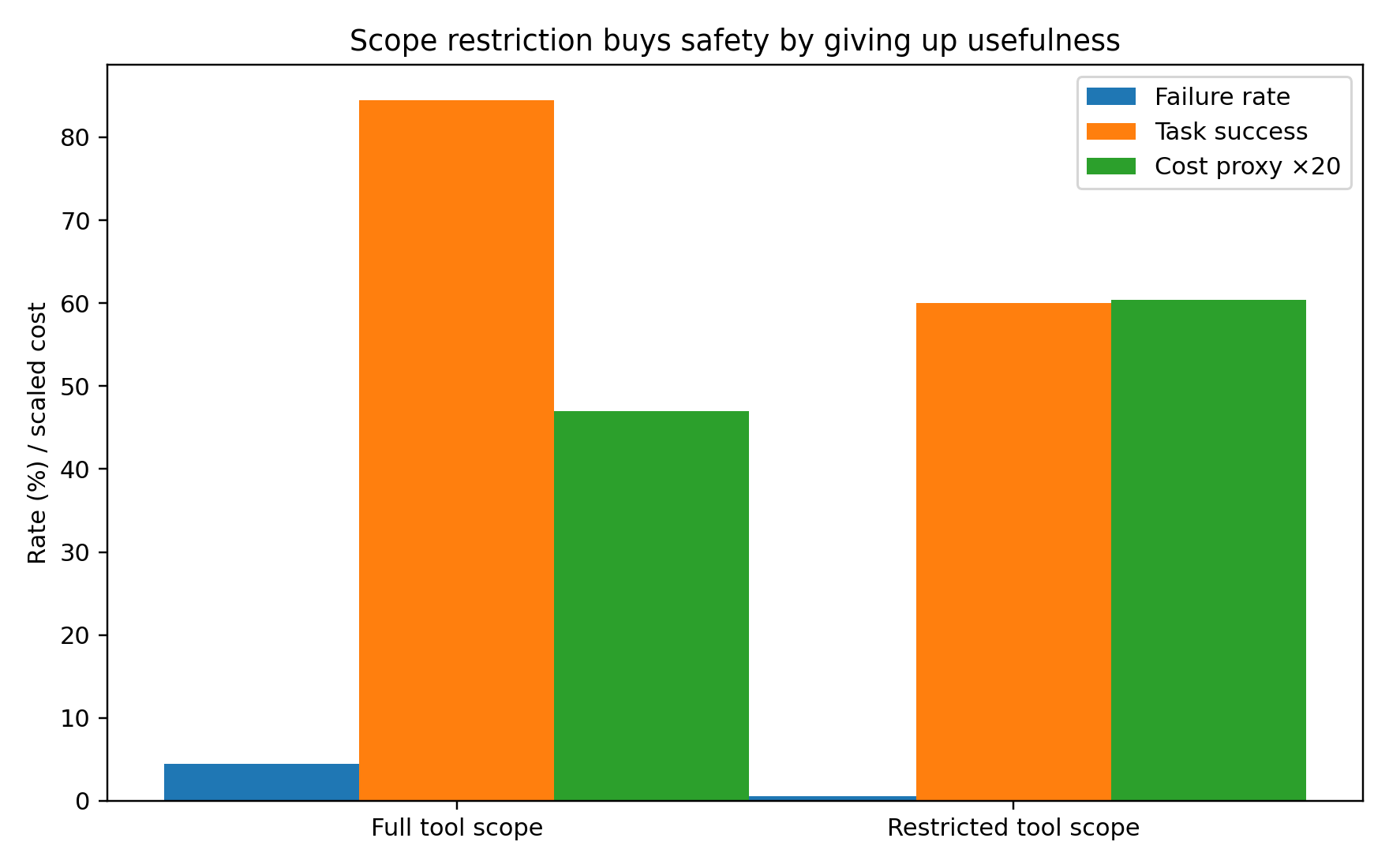}
\caption{Scope restriction buys safety by giving up usefulness.}
\figurenote{The figure compares full and restricted tool scope. Restricted scope sharply lowers catastrophic failure but also reduces task success. When access is insufficient, the deployer may substitute by narrowing the system's permitted actions.}
\end{figure}

\vspace{0.5\baselineskip}
\subsection{Sovereignty discount: cost-shifting, not just cost increase}\label{interpreting-the-sovereignty-discount-proxy}

The sovereignty-discount analysis shows that bounded sovereignty is best
understood as \textbf{cost shifting}, not simply cost increase. Some
bounded-access profiles can reach loose safety targets at lower direct cost
than the full-instrumentation baseline, especially by restricting scope or
relying on a benign agent. That does not mean bounded sovereignty is cheaper;
it means the cost has moved into another dimension, such as reduced
usefulness, weaker diagnosis, delayed vendor assurance, residual incident
risk, or greater manual effort.

The stricter safety-plus-usefulness target is therefore more informative
than a simple cost-to-safety measure. Under a $\leq$1\% failure and $\geq$75\%
usefulness target, only \textbf{35\%} of access profiles were reachable
against the stealthy access-aware agent and \textbf{40\%} against the
careless agent. For readers concerned with real-world governance, the key
question is not only ``Can failure be reduced?'' but ``Can failure be reduced
while preserving enough useful automation and enough evidence for diagnosis?''

\minorheading{Summary of empirical contribution}\label{summary-of-empirical-contribution}

The synthetic experiment supports four modest claims. First,
control-protocol feasibility is access-dependent: the same protocol behaves
differently depending on whether the deployer has logs, tool visibility,
gateway control, trace access, model-version control, and vendor support.
Second, complete logging and gateway ownership are the most important
interaction-layer controls in this setting because logs improve monitoring
and diagnosis while the gateway enables prevention. Third, model-side access
matters most for diagnosis and assurance, because trace access and
model-version control improve reconstructability even when output-level
monitoring remains possible. Fourth, scope restriction is a real substitute
for missing access, but it improves safety by reducing what the system can do.
These findings do not measure real payment systems; they show that access
assumptions are empirically meaningful features of control-protocol
feasibility.

\section{Illustrative Case Interpretation: National Payments
Infrastructure}\label{illustrative-case-interpretation-national-payments-infrastructure}

This section is an illustrative and anonymised interpretation, not a
report on a specific real deployment or incident. It exists to map the
conceptual and synthetic findings to the class of deployments the paper
is concerned with: regulated infrastructure operators in AI-importing
jurisdictions using vendor-supplied foundation models.

\subsection{Illustrative deployment and access audit}\label{deployment-description}

Consider a national payments infrastructure operator in an AI-importing
jurisdiction. The national financial regulator supervises the Operator,
which runs interbank retail payment rails. The regulator's model risk
management expectations require documented validation, ongoing
monitoring, auditability, and board accountability for material models.

The Operator deploys a vendor-supplied frontier foundation model through
a managed cloud endpoint in a foreign-operated cloud region with a
data-residency rider. The model is used in an agentic,
transaction-adjacent function: for example, an assistant that triages
disputed transactions, drafts investigation notes, and invokes a
constrained set of internal tools such as case lookup, recommendation
drafting, and escalation.

The catastrophic-action classes, in ascending severity, are:

\begin{itemize}
\tightlist
\item
  \textbf{C1:} an incorrect hold or release on an individual
  transaction;
\item
  \textbf{C2:} systematic mis-triage that propagates across many cases
  before detection;
\item
  \textbf{C3:} exposure of protected personal financial data through
  tool calls or outputs; and
\item
  \textbf{C4:} any action path that could disturb settlement operations.
\end{itemize}

\clearpage
\minorheading{Access audit}\label{access-audit}

Applying the four-layer typology produces the following access audit.

\begin{table}[H]
\centering
\small
\setlength{\tabcolsep}{4pt}
\renewcommand{\arraystretch}{0.95}
\begin{tabular}{@{}
  >{\raggedright\arraybackslash}p{0.20\linewidth}
  >{\raggedright\arraybackslash}p{0.20\linewidth}
  >{\raggedright\arraybackslash}p{0.52\linewidth}@{}}
\toprule
Layer & Operator's position & Basis \\
\midrule
Data & Partial & Full control of own case and transaction data used in prompts or fine-tuning; no visibility into base-model training data \\
Model & None & Managed-endpoint access; no weights, architecture detail, activations, or native patching ability; vendor may update the served model \\
Infrastructure & Partial & Data-residency rider constrains storage location; serving stack remains vendor-operated and not independently observable; legal reach is mediated by foreign law \\
Interaction & Partial & I/O logging and tool-call logging may be contractable and implemented at the Operator's gateway; internal traces remain unavailable \\
\bottomrule
\end{tabular}
\end{table}

This is not an unusual position for a regulated AI importer: the
deployer may control the business process and the local tool surface but
not the model, weights, infrastructure, internal traces, or update
process.

\subsection{Protocol feasibility under bounded sovereignty}\label{protocol-feasibility-mapping}

Reading the access audit against the protocol requirements matrix,
output-level trusted monitoring is executable if a smaller,
Operator-controlled monitor scores every output and tool call on
Operator-controlled compute. Human auditing of the highest-suspicion
fraction is also executable, subject to audit capacity and log
completeness.

Defer-to-trusted routing is executable if and only if the agent's tool
calls pass through an Operator-owned execution gateway where they can be
held, scored, replaced, or escalated before execution. This is the
central architectural lesson from the synthetic study: the gateway is
the point at which detection becomes prevention.

Resampling-style protocols are degraded. API re-querying is possible,
but vendor model-version changes can weaken reproducibility; per-query
cost and latency bind; and resampling does not substitute for missing
trace access. Pre-deployment control evaluations are also degraded
because they evaluate whichever model version the API serves at
evaluation time unless the deployer can pin the production model
version.

Anything requiring internal traces is infeasible without vendor
cooperation: activation-level monitoring, probe-based deception
detection, and interpretability-supported auditing. This matters because
C2- and C3-class failures with a deceptive or systemic character are
precisely where output-only monitoring is weakest in principle.

\subsection{Sovereignty discount cost and stress scenario}\label{where-the-sovereignty-discount-cost-concentrates}

The case interpretation locates sovereignty discount cost in four
places.

First, \textbf{contractual spend}: negotiating complete I/O logging,
model-version change notification, audit and attestation rights,
incident-notification clauses, and regulator-access provisions that a
weight-holding deployer would not need to buy.

Second, \textbf{architectural spend}: building and operating the
execution gateway, tamper-evident logs, deployer-side monitors, and
controlled tool interfaces.

Third, \textbf{capacity spend}: staffing audit and model-risk functions
able to consume monitor outputs, investigate suspicious patterns, and
challenge vendor attestations.

Fourth, \textbf{scope-restriction cost}: the usefulness forgone by
excluding C4-adjacent action paths and constraining the tool set when
monitoring cannot reach deeply enough.

The synthetic study gives this interpretation empirical shape. It shows
that gateway access reduces synthetic failure and increases intervention
success; complete logs improve diagnosis; trace and version access
strengthen post-incident explanation; and scope restriction lowers
failure by reducing usefulness.

\minorheading{Illustrative stress scenario}\label{illustrative-stress-scenario}

Suppose the monitor flags a slow elevation in hold-placement
recommendations affecting one merchant category. Under full access, the
natural diagnostic would include inspecting internal traces, re-running
a pinned model version on historical cases, comparing version behaviour,
and reconstructing the full tool-call chain.

Under the Operator's bounded access, the diagnostic set is narrower:
output-log forensics, tool-call review if the gateway captured it,
re-querying a model that may already have changed, and vendor support
escalation governed by contract. The deployer's detection may survive
bounded sovereignty, but diagnosis and intervention are the first
casualties. The difference is invisible until an incident makes it
visible.

\section{Discussion and Limitations}\label{discussion-and-limitations}

\subsection{What bounded control can and cannot do}\label{revisiting-the-three-positions}

The conceptual framework, synthetic experiment, and illustrative case do
not settle the absolute-versus-bounded-control question. They support a
narrower claim.

Against a strong reading of Position A, bounded control is not empty. A
non-trivial subset of canonical control protocols is executable under
partial sovereignty, provided the deployer secures two load-bearing
elements: complete interaction logging and an owned execution gateway.

Against a strong reading of Position B, full-stack sovereignty is not
the only route to some safety-relevant assurance. Architecture and
contract can do meaningful work, especially for output-level monitoring,
gateway-based intervention, and human audit.

In support of both A and B, the infeasible set is not marginal. It
contains precisely the internal-access protocols most relevant to
deceptive, systemic, or hard-to-diagnose failure modes. Contractual
substitutes are also revocable or incomplete in ways native access is
not. The honest summary is that bounded control is real but has a
ceiling, and the ceiling is lowest for the failure modes that most worry
the control literature.

\minorheading{What the empirical study adds}\label{what-the-empirical-study-adds}

The synthetic access-ablation study strengthens the paper in one
specific way: it turns the access typology into variables that can be
manipulated. The results show that access assumptions are not background
implementation details. They affect prevention, diagnosis, usefulness,
cost, and target reachability.

The most important practical distinction is between \textbf{seeing} and
\textbf{stopping}. Logging improves visibility and diagnosis, but a
gateway enables intervention. The second distinction is between
\textbf{detecting} and \textbf{explaining}. Output monitoring may flag a
risky action, but trace access, model-version control, and vendor
support affect whether the deployer can reconstruct why it happened. The
third distinction is between \textbf{buying safety} and \textbf{giving
up usefulness}. Scope restriction can reduce synthetic failure, but it
does so by reducing what the system can do.

These distinctions are useful for AI safety research because they make
access assumptions explicit. They are useful for governance readers
because they translate abstract sovereignty into practical questions: Do
we have complete logs? Can we intercept tool calls before execution? Can
we pin the model version? Can we reconstruct failures? Can the regulator
access the evidence needed to assess the safety case?

\subsection{Generalisability and limitations}\label{generalisability}

The mapping method is sector-agnostic; the case interpretation is not.
Payments infrastructure is favourable terrain for bounded control
because catastrophic actions are relatively crisp, tool surfaces can be
narrowed, and a gateway architecture is natural. Healthcare, public
administration, or general enterprise deployments may have fuzzier
definitions of catastrophe and wider action surfaces. That would likely
raise the importance of trace access, richer audit evidence, and scope
restriction.

The empirical workflow is also synthetic and is designed to test the
logic of access constraints, not to predict failure rates in any real
system. A second sector case and an LLM-in-the-loop version of the
synthetic environment would strengthen external validity.

\minorheading{Limitations}\label{limitations}

Six limitations matter.

First, the empirical study is synthetic. It does not use real payment
data, real customer data, production logs, vendor contracts,
confidential institutional material, or operational payment
infrastructure. It is a structured simulation of the framework, not
measured evidence about any actual institution.

Second, the synthetic agents are not frontier models. They are useful
for separating honest, careless, and access-aware behaviour, but they do
not capture the full ambiguity of real model outputs. A future
LLM-in-the-loop version would be needed to test whether real model
outputs create different monitor calibration, audit burden, or diagnosis
patterns.

Third, the access profiles are a curated fractional-factorial design.
This is appropriate for isolating the paper's main access factors, but
it does not cover every possible combination of cloud, API, contractual,
logging, data-residency, and vendor-support arrangements. A
full-factorial run is supported by the code but would be computationally
larger.

Fourth, the cost model is a proxy. It decomposes inference, monitoring,
audit, logging, gateway, vendor assurance, latency, scope restriction,
and residual incident loss, but it is not a financial estimate.
Accordingly, the results support relative cost-proxy comparisons rather
than exact control-tax or sovereignty-discount amounts.

Fifth, the sovereignty-discount proxy is best read as a reachability and
cost-shifting measure. In some synthetic cases, bounded-access profiles
can reach loose safety targets at lower direct cost than the
full-instrumentation baseline because they restrict scope or accept
weaker usefulness and diagnosis. This is not evidence that bounded
sovereignty is cheaper; it shows that missing access can shift cost into
other dimensions.

Sixth, the literature underpinning sovereign AI is fast-moving and
partly non-peer-reviewed. The four-pillar sovereignty formulation is a
policy and industry construct that this paper adapts rather than a
settled academic taxonomy. The paper therefore frames its contribution
narrowly: it does not claim to settle the meaning of AI sovereignty, but
shows how access constraints can be specified, tested, and priced within
deployment-time AI control.

\subsection{Future work}\label{what-would-most-strengthen-the-paper-next}

The next research step is external grounding, and three extensions would be
especially valuable.

First, an LLM-in-the-loop synthetic workflow would test whether real
model outputs create different ambiguity, monitoring challenges, and
audit burdens than scripted synthetic agents.

Second, a public vendor and regulatory access-rights audit would convert
the access typology into an evidence table. It could ask which vendors
offer model-version pinning, log export, tool-call traces,
data-retention control, customer-managed keys, regulator access,
incident forensics, and vendor attestations.

Third, expert calibration would improve the cost model. Practitioners
could provide ranges for audit capacity, model-version drift, vendor
response delay, log-retention cost, operational latency cost, and
acceptable failure thresholds without disclosing confidential
information.

\subsection{Scope boundary}\label{scope-boundary}

This paper makes no export-control, trade-policy, procurement-policy, or
industrial-policy recommendation. It takes no position on whether
jurisdictions should pursue sovereign AI capability; those questions
belong to a separate policy-facing analysis with a different evidentiary
standard. This paper's recommendation is narrower: control-protocol
specifications should state their access assumptions explicitly.

\section{Conclusion}\label{conclusion}

AI control research has begun to price oversight through concepts such as
the control tax. This paper argues that oversight must also be located. A
control protocol is not only a technical design; it is also an access claim.
It assumes that someone can observe the right logs, route the right outputs,
intercept the right tool calls, pin or identify the right model version, and
reconstruct the right evidence after an incident. For many regulated
deployers, especially in AI-importing jurisdictions, these assumptions do not
fully hold.

The paper introduced \textbf{bounded sovereignty} to describe this condition:
partial access across data, model, infrastructure, and interaction layers. It
then provided a protocol-by-layer requirements matrix for asking which
control protocols remain feasible under partial access, and introduced
\textbf{sovereignty discount cost} as the part of the control tax paid when a
deployer must substitute for missing access through contracts, architecture,
audit, vendor assurance, residual risk, or scope restriction.

The synthetic access-ablation experiment over 1.35 million synthetic case
simulations made the argument concrete. Gateway access reduced failure and
enabled intervention. Complete logs improved diagnosis. Trace access and
model-version control strengthened post-incident explanation. Scope
restriction improved safety, but by reducing usefulness. These results are
not evidence about real payment systems; they show that bounded sovereignty
can be operationalised and that access constraints change the
safety-usefulness-cost frontier.

The practical recommendation is simple. AI-control papers should state their
access assumptions explicitly. A protocol that assumes trace access,
model-version pinning, tool-call interception, complete interaction logs, or
vendor cooperation should say so. For policymakers, technocrats, and
regulated deployers, those assumptions are not implementation details. They
are often the binding constraint.

\end{document}